\documentclass[runningheads]{llncs}

\usepackage[mobile]{eccv}

\usepackage{eccvabbrv}

\usepackage{graphicx}
\usepackage{booktabs}

\newcommand\blfootnote[1]{% 
\begingroup 
\renewcommand\thefootnote{}\footnote{#1}% 
\addtocounter{footnote}{-1}% 
\endgroup 
}

\usepackage[accsupp]{axessibility}  % Improves PDF readability for those with disabilities.

\usepackage[pagebackref,breaklinks,colorlinks,citecolor=eccvblue]{hyperref}
\usepackage{orcidlink}

\usepackage{ulem}
\usepackage{multirow}
\usepackage{makecell}

\usepackage{algorithm}
\usepackage[noend]{algpseudocode}
\usepackage{algorithmicx}
\usepackage{xcolor}
\usepackage{colortbl}
\usepackage{makecell}
\usepackage{amsmath}
\usepackage{tabularx} % Remember to include this line in the preamble
\usepackage{booktabs} % For professional looking tables
\begin{document}

\title{AnimatableDreamer: Text-Guided Non-rigid 3D Model Generation and Reconstruction with Canonical Score Distillation} 

\titlerunning{AnimatableDreamer}

% Second Author\inst{2,3}\orcidlink{1111-2222-3333-4444} \and
% Third Author\inst{3}\orcidlink{2222--3333-4444-5555}}

% % TODO FINAL: Replace with an abbreviated list of authors.

% % TODO FINAL: Replace with your institution list.
% Springer Heidelberg, Tiergartenstr.~17, 69121 Heidelberg, Germany

\author{Xinzhou Wang\inst{1,2,3} \and
Yikai Wang\inst{2}\textsuperscript{\dag} \and
Junliang Ye\inst{2} \and
Fuchun Sun\inst{2}\textsuperscript{\dag} \and
Zhengyi Wang\inst{2,3} \and
Ling Wang\inst{2,5} \and
Pengkun Liu\inst{2,4} \and
Kai Sun\inst{2} \and
Xintong Wang\inst{6} \and
Wende Xie\inst{7} \and
Fangfu Liu\inst{2} \and
Bin He\inst{1}}

\authorrunning{X. Wang et al.}

\institute{
$^1$ Tongji University \quad $^2$ Tsinghua University \quad $^3$ ShengShu \quad $^4$ Fudan University \\ $^5$ Xi'an Research Institute of High-Tech \quad $^6$ Zhejiang University \quad $^7$ Didi}
% \institute{Tongji University, Shanghai, China and $^2$ Beijing National Research Center for Information Science and Technology (BNRist), State Key Lab on Intelligent Technology and Systems, Department of Computer Science and Technology, Tsinghua University, Beijing, China \quad $^3$ ShengShu, Beijing, China \\
% $^4$ Academy for Engineering and Technology, Fudan University, Shanghai, China \quad $^5$ Xi'an Research Institute of High-Tech, China \quad $^6$
% Zhejiang University, China}

\maketitle
\blfootnote{$^\dag$ Corresponding authors.} 

\begin{abstract}
\vspace{-1.3em}
Advances in 3D generation have facilitated sequential 3D model generation (a.k.a 4D generation), yet its application for animatable objects with large motion remains scarce. Our work proposes {AnimatableDreamer}, a text-to-4D generation framework capable of generating diverse categories of non-rigid objects on skeletons extracted from a monocular video. At its core, {AnimatableDreamer} is equipped with our novel optimization design dubbed Canonical Score Distillation (CSD), which lifts 2D diffusion for temporal consistent 4D generation. 
CSD, designed from a score gradient perspective, generates a canonical model with warp-robustness across different articulations. Notably, it also enhances the authenticity of bones and skinning by integrating inductive priors from a diffusion model. Furthermore, with multi-view distillation, CSD infers invisible regions, thereby improving the fidelity of monocular non-rigid reconstruction. Extensive experiments demonstrate the capability of our method in generating high-flexibility text-guided 3D models from the monocular video, while also showing improved reconstruction performance over existing non-rigid reconstruction methods. 

\textbf{Project page} \url{https://zz7379.github.io/AnimatableDreamer/}.

\end{abstract}
\vspace{-2em}

\begin{figure*}[t]
\vspace{-0.6em}
  \centering
   \includegraphics[width=1\linewidth]{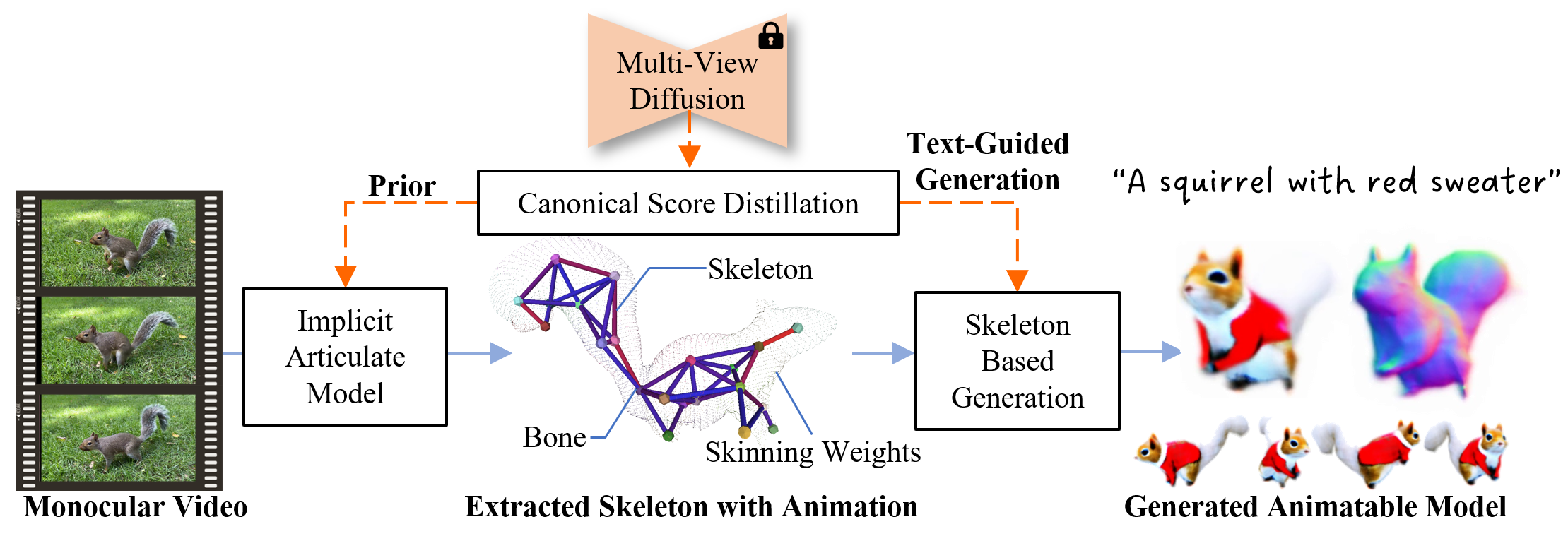}
   \vspace{-1.6 em}
   \caption{Given a monocular generic category video, AnimatabaleDreamer initially extracts the skeleton with skinning and motions assisted byprior from a diffusion model. Subsequently, through our canonical score distillation, AnimatabaleDreamer generates novel animatable 3D models from the extracted skeleton and a text prompt with temporal consistency and warping robustness.}
   \label{fig:teaser}
   \vspace{-1.5 em}
\end{figure*}
\vspace{-1 em}
\section{Introduction}
\label{sec:intro}
Automatically building animatable 3D models with non-rigid deformations and motions plays a crucial role in broad fields such as gaming, virtual reality, film special effects, etc. With the remarkable success of deep generative models, generating various 2D images through text prompts comes to reality~\cite{zhang2023adding, rombach2022high, saharia2022photorealistic}, and this success is expanding beyond 2D generation. The application of Score Distillation Sampling (SDS)~\cite{poole2022dreamfusion} has elevated 2D text-to-image diffusion models to generate high-quality 3D models. Numerous subsequent works have emerged in this domain~\cite{lin2023magic3d, shi2023mvdream, wang2023prolificdreamer, tang2023dreamgaussian}. However, generating deformable objects remains challenging due to their inherent unconstrained and ill-conditioned nature~\cite{park2021nerfies, park2021hypernerf, wang2022fourier}. 

Intuitively, bones and motions extracted by implicit animatable models could serve as geometry constraints for deformable object generation. Recent efforts have been devoted to reconstructing animatable 3D models with pre-defined or learned skeletons~\cite{stathopoulos2023learning, yang2023reconstructing, wu2023magicpony, jakab2023farm3d, yao2023artic3d}. Nevertheless, these methods are mostly category-specific with limited diversity or largely rest on the captured multi-view data~\cite{yang2022banmo,su2021nerf, cheng2023dna}. We believe this problem can be greatly alleviated by distilling 2D priors from the diffusion model to hallucinate plausible geometry of invisible regions and avoid short-cut solutions~\cite{niemeyer2022regnerf}.

% We believe there exists a synergy between animatable model generation and reconstruction, which potentially promotes performance for one another. 

To address these challenges, we propose \textbf{AnimatableDreamer}, a two-stage framework designed to extract skeletons from monocular videos and generate generic categories of non-rigid 3D models on these skeletons. Initially, the non-rigid object in the monocular video is disentangled into a canonical implicit field~\cite{chen2021snarf, mildenhall2021nerf} with a skeleton-based structure consisting of bones and neural skinning. AnimatableDreamer extracts bones and skinning from a monocular video, leveraging multi-view diffusion priors to refine the warping, geometry, and texture of unseen regions. Skeletons are generated based on the skinning weights of vertices and further constrain the pairwise relationship between bones of the generated model. 
Subsequently, under the constraint of the extracted skeleton and a specific text prompt, AnimatableDreamer generates 4D content with a diffusion model. Considering that directly employing SDS\cite{poole2022dreamfusion} or Variational Score Distillation (VSD)\cite{wang2023prolificdreamer} will make the canonical model detach from the extracted motions and harm the plausibility of the warped model, we propose \textbf{Canonical Score Distillation (CSD)} to generate novel non-rigid 3D models. CSD is a novel distilling strategy designed to simultaneously generate a canonical model aligned with motions and refine the skeletons and skinning. CSD denoises multiple warped models through invertible warping functions while consistently optimizing a static canonical space shared by all animation frames. This novel approach simplifies 4D generation into a more manageable 3D process, yet maintains comprehensive supervision throughout the 4D space and ensures the morphological plausibility of the model under various object poses. Furthermore, CSD refines the motions and skinning weights to ensure consistency with the canonical model.
% \textbf{Canonical Score Distillation (CSD)}, a novel distilling strategy, is design to attain the synergy of both animatable generation and reconstruction. Through invertible warping functions, CSD denoises multiple camera spaces while consistently optimizing a static canonical space shared by all video frames. This novel approach simplifies 4D generation into a more manageable 3D process, yet maintains comprehensive supervision throughout the 4D space and ensures the morphological plausibility of the model under various object poses.

% 	AD：以diffusion作为先验，从单目视频中提取关节及运动，生成骨架，并在得到的骨架上进行4D生成。
% 	提出CSD，CSD可利用diffusion优化skel及生成模型几何和纹理。通过XXX相机空间到同一个规范空间，CSD保证了4D生成的可靠性。
% 	实验结果：利用该方式生成的骨架进行重建XXX，生成效果比其他的也sota

% AD 从单目视频中提取bone及运动，生成skeleton，并用于4D生成
% CSD 对重建生成都有好处，其两项分别
% 第一个有skeleton约束的3D生成，利用skinning weight 生成skeleton，并利用skeleton和bone约束CSD生成

To summarise, we make the following contributions:
\begin{itemize}
    \item \textbf{{AnimatableDreamer}}: A novel framework that extracts skeletons with motions from a monocular video and generates generic categories of non-rigid 3D models based on these skeletons. This is the first implementation of generating text-guided non-rigid 4D content leveraging video-based skeletons.
    
        \item \textbf{Canonical Score Distillation}: A new distillation method enhances the generation and reconstruction of non-rigid 3D models. By back-propagating gradients from multiple camera spaces to a static canonical space, CSD ensures the morphological plausibility of models after warping. Besides, CSD refines bones and skinning weights through a specifically tailored gradient term. With these designs, CSD improves the reconstruction quality of unseen regions with diffusion prior and is capable of generating 4D models with time consistency and warping robustness.
    
    \item \textbf{Skeleton-based Generation}: An innovative approach for 4D generation taking skeletons, bones, skinning weights, and motions as prior. With constructed skeletons,  a constraint with $SE(3)$  is utilized to guide the transformations of bone pairs, thereby preventing motion detaching and ensuring convergence. Furthermore, the density of Gaussian bones is considered as an indicator for the generation of surfaces with warping robustness.
\end{itemize}
\vspace{-1 em}
\section{Related Work}
\label{sec:relatedwork}
\vspace{-0.5 em}
\subsection{Neural Reconstruction for 3D Non-rigid Object}
Neural Radiance Field (NeRF) has been a groundbreaking advancement in representing static scenes, enabling the generation of photorealistic novel views and detailed geometry reconstruction~\cite{mildenhall2021nerf, barron2021mip, barron2022mip, li2023neuralangelo}. Adapting NeRF for dynamic scenarios has involved augmenting the field into higher dimensions to accommodate objects with changing topologies~\cite{park2021hypernerf}. An alternative strategy in dynamic object reconstruction employs an additional warping field to deform the NeRF~\cite{park2021nerfies, pumarola_d-nerf_2021}. Nonetheless, these dynamic NeRF adaptations often encounter performance degradation, primarily due to the complexities introduced by the added temporal dimension. This could be alleviated by applying alternative representations including tesnors~\cite{shao_tensor4d_2023}, Gaussian Splatting~\cite{wu_4d_2023} and explicit representation~\cite{cao2023hexplane, fridovich2023k}. Despite these innovations, synthesizing space-time views from monocular perspectives remains a significant hurdle. Implementing spatio-temporal regularization methods, including depth and flow regularization, has shown potential in overcoming this issue~\cite{du2021neural, xian2021space}. Furthermore, exploring category-specific or articulate priors offers promising avenues for reconstructing non-rigid objects~\cite{peng2021neural, wang2023rpd, cheng2023dna, yang2022banmo, yang2023reconstructing, Yang_2023_ICCV}. These approaches offer novel opportunities and insights for advancing the field of 3D non-rigid object reconstruction. However, they often overlook the application of generic priors, which could reduce the reliance on domain-specific priors and manually designed templates. Contrarily, our method leverages model training on extensive, generic datasets to distill such priors, thereby enhancing the reconstruction process.
\vspace{-1 em}
\subsection{Distillation-based 3D Generation from Diffusion Model}
SDS~\cite{poole2022dreamfusion} has gained prominence for its capability to elevate pre-trained 2D diffusion models to the realm of 3D generation. By distillate 2D prior learned from large-scale datasets and optimizing implicit field~\cite{mildenhall2021nerf}, SDS is able to generate high-quality 3D models based on text-prompt~\cite{wang_score_2022,chen_scenedreamer_2023,wang2023prolificdreamer}. The integration of differentiable marching tetrahedra~\cite{shen2021dmtet} further enhances the combination of explicit meshes and SDS~\cite{lin2023magic3d}. However, semantic consistency challenges arise in distillation-based 3D generation methods due to their disconnection from the 3D dataset during training. Addressing this concern, MVDream~\cite{shi2023mvdream} introduces a multi-view diffusion model for panoramas with homography-guided attention, improving semantic consistency by incorporating cross-view attention and camera conditions. Further, 2D diffusion model trained can be lifted to 4D via a temporal score distillation sampling~\cite{singer_text--4d_2023}, which integrates world knowledge into 3D temporal representations. In contrast, our method aims to produce a time-consistent and warp-robust 4D model by initially generating a non-rigid model based on a skeleton extracted from the video. By applying this model across various animations, we ensure morphological plausibility even when the skeletons exhibit differing articulations. 

% \subsection{Category-specific animatable models}

\vspace{-1 em}
\section{Method}
\label{sec:method}
\begin{figure*}[ht]
\vspace{-0.3em}
  \centering
   \includegraphics[width=0.95\linewidth]{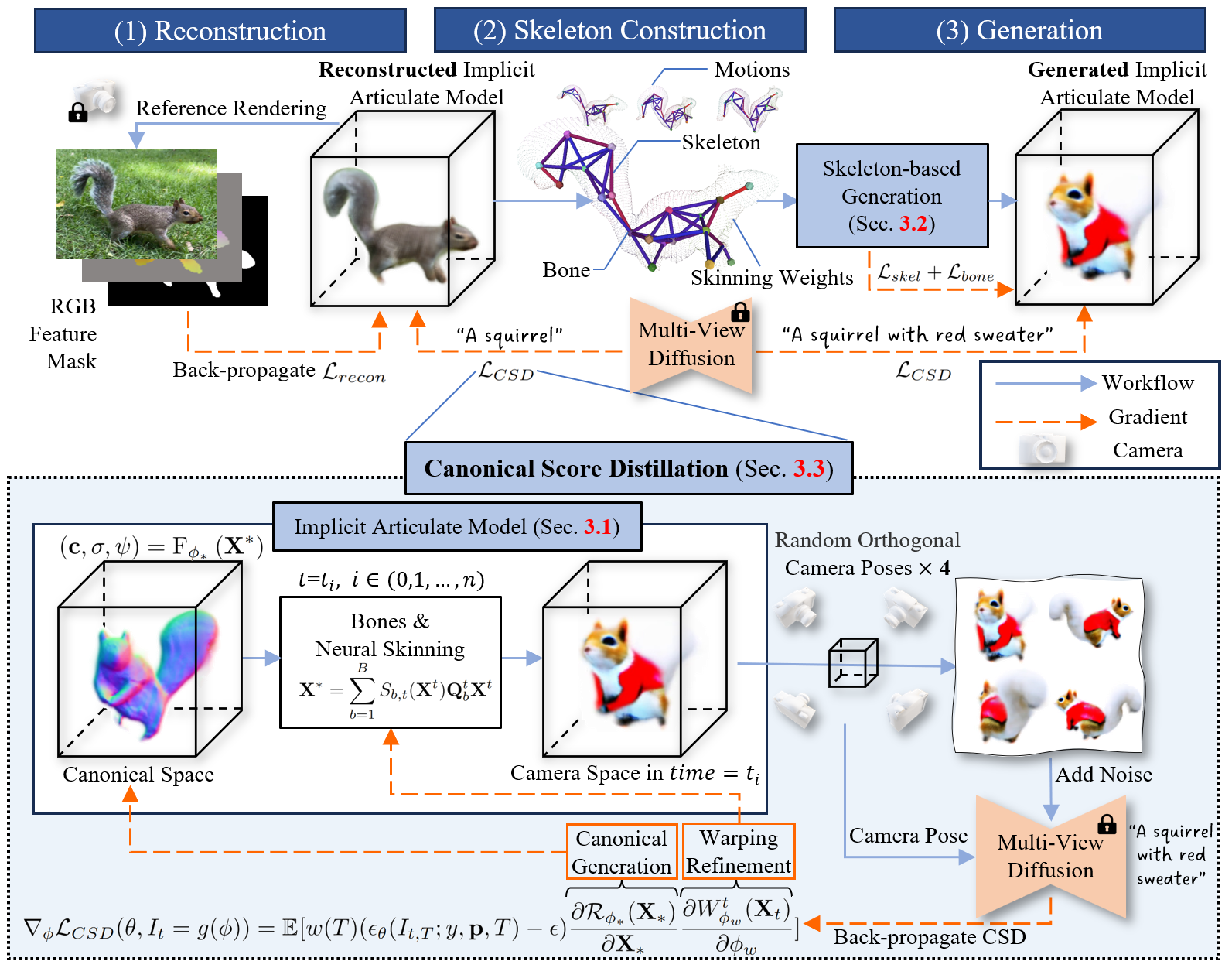}
   \vspace{-0.3 em}
   \caption{\textbf{Framework overview.} 
\textbf{Top:} The AnimatableDreamer framework extracts the skeletons, skinning, and animation from monocular generic category videos with CSD to enhance the invisible regions. Subsequently, through our canonical distillation strategy, AnimatabaleDreamer generates text-guided novel animatable 3D models on the extracted skeleton. \textbf{Bottom:} We decompose the articulated model into a static neural field and time-varying neural skinning that transforms the object from canonical space to camera space. During training, we rendered a warped model on four random orthogonal views and optimized the model in camera space across different frames. The canonical generation term in CSD enhances the morphological plausibility of warped models, while the warping refinement term further refines the bone motions and skinning. }
   \label{fig:overview}
   \vspace{-2 em}
\end{figure*}

\vspace{-0.5 em}
Given a monocular video $V = \{(I_i, t_i)\}_{i=1}^{n}$, our objective is two-fold: first, to generate an object related to a specified prompt $\textbf{y}$ on the skeletons and rigging extracted from the provided video; second, to reconstruct the original object with diffusion prior. % We base our articulated model on the framework of BANMo~\cite{yang2022banmo} with some modifications (\cref{sec:model}) and explore its potential in deformable object generation. 
The proposed framework, AnimatableDreamer, as illustrated in \cref{fig:overview}, comprises two distinct stages: skeletons extraction (\cref{sec:model}) and skeletons-based generation (\cref{sec:skeleton}). Both stages employ CSD for content generation and warping refinement (\cref{sec:csd}). These workflows supervise the deformed model in \textbf{camera space} (articulated poses) and optimize the model in \textbf{canonical space} (rest pose) through differentiable warping.

\vspace{-1 em}
\vspace{-0.3em}
\subsection{Implicit Articulate Model}
\label{sec:model}
% We build our representation based on BANMo\cite{yang2022banmo} mainly for two considerations: First, to empower the framework with ability of generation, the model should be flexible enough for topological changes. Most previous work employed explicit meshes[Learning category-specific mesh reconstruction from image collections, LASR: Learning articulated shape reconstruction from a monocular video, DOVE: Learning deformable 3d objects by watching videos] or combine the implicit field with deformable mesh\cite{shen2021dmtet}[fram3d magicpony]. Rendering mesh with differentiable rasterization\cite{nvdiffrast} enable gradient to back-propagate but compared to volume rendering, differentiable rasterization can only optimize a few region in the field, which limit its ability of topological deformation. Second, it is a highly unconstrained problem to learn articulate model from single monocular video without template. Thus the motion modeling should be compact enough for recovering deformation from limited visual cue. Though flexible and of high quality, previous dense motion field[nfss, d-nerf,...] need multi-view supervising to rebuilt the motion faithfully. On contrary to motion and model representation mentioned above, BANMo build implicit field with neural blend skinning, which inherently constrain the continuity of motion field both temporally and spatially. 
\textbf{Canonical Model.} We utilize the NeuS model~\cite{wang2021neus} as our canonical representation to accurately reconstruct surface geometries, in conjunction with rendering an additional feature descriptor~\cite{yang2021viser} to incorporate priors from off-the-shelf methods~\cite{li2021neural,dinov21} for self-supervised 3D registration. This approach facilitates the articulation extraction and 3D reconstruction of objects across various categories from monocular video. In our representation, each 3D point $\mathbf{X} \in \mathbb{R}^3$ on the canonical model is characterized by a color vector $\mathbf{c}\in \mathbb{R}^3$, a Signed Distance Field (SDF) value $\mathbf{d}\in \mathbb{R}$, and a feature descriptor $\psi \in \mathbb{R}^{16}$:
\begin{align}
    (\mathbf{c,d})&=\mathbf{MLP}_*\left(\mathbf{X}, \mathbf{v}\right),\\
    \psi&=\mathbf{MLP}_\psi \left(\mathbf{X}\right),
    \label{eq:nerf}
\end{align}
where $\mathbf{v} \in SO(3)$ is view direction. Through the cumulative function of an unimodal distribution $\Gamma_{\beta}(\cdot)$, the Signed Distance Function (SDF) is transformed into a density representation~\cite{wang2023neus2,wang2021neus}. The feature descriptor $\psi$ is learned based on 2D features extracted from the self-supervised vision model, DINOV2~\cite{dinov21}. Considering that CSD supervises a point along the time axis without strong texture continuity constraints, we employ a time-invariant canonical model to prevent texture flickering. 
%-------------------------------------------

\noindent\textbf{Warping Field.} In contrast to dense motion fields~\cite{park2021nerfies, park2021hypernerf, cao2023hexplane}, we disentangle non-rigid objects into a canonical model and a compact motion field~\cite{yang2022banmo, noguchi2022watch}. Such disentanglement mitigates the challenges associated with the ill-conditioned nature inherent to 4D generation and enables the application of distillation-based 3D generators. To warp the field from camera space to canonical space, we build the mapping between the 3D point in canonical space $\mathbf{X}^*$ and 3D point in camera space $\mathbf{X}^t$ through a blend skinning deformation defined on $\textit{B}$ rigid bones:
\vspace{-0.5em}
\begin{align}
   \mathbf{X}^{*}={W}_{\phi_w}(\mathbf{X}^{t}) &= \sum_{b=1}^{B} S_{b,t}(\mathbf{X}^t) \mathbf{Q}_b^t \mathbf{X}^t, \label{eq:wb}
\end{align}
where ${W}(\cdot)$ is warping field, $\phi_w$ are warping-related parameters, and transformations of the $b$-th bone $\mathbf{Q}_b^t\in SE(3)$ is dual quaternion blend skinning (DQB) ~\cite{kavan2007dqb} learned from $\mathbf{MLP_Q}(t)$. Here we applied the Fourier function for time embedding ~\cite{mildenhall2021nerf}. $S_{b,t}(X^t)$ is skinning weight of the $b$-th bone related to $X^t$:
\begin{align}
   S_{b,t}(\mathbf{X}^t) &= \mathcal{M}(\mathbf{Q}_b^t, \mathbf{\sigma}_b, \mathbf{X}^t) + \Delta S_{b,t}. \label{eq:skin}
\end{align}
Here $\mathcal{M}$ is the Mahalanobis distance between $\mathbf{X}^t$ and the $b$-th bone at time $t$. Each Gaussian bone is assigned with a learnable scaling parameter $\mathbf{\sigma}_b$. The term $\Delta S_{b,t}$ represents the delta skinning weights derived from $\mathbf{MLP}_\Delta$. It is important to note that the global motion of the object is integrated into the camera poses to enhance clarity. The warping-related parameters $\phi_w$ includes $\mathbf{MLP}_Q$, $\mathbf{\sigma}_b$, and $\mathbf{MLP}_\Delta$. To render a pixel $\mathbf{c}$ in camera space for a given time $t$, we warp camera space sampling points $\mathbf{X^t_i} \in \mathbb{R}^3$ to canonical space with warping function and apply volume rendering:
\vspace{-0.5 em}
\begin{align}
    \mathbf{c}\left(\mathbf{x}^{t}\right)=\mathcal{R}_{\phi_*}(\mathbf{X}^{*}(\mathbf{X}^{t})), \label{eq:volumerendering}
\end{align}
where $\phi_*$ are parameters of $\mathbf{MLP}_*$ defined in \cref{eq:nerf}. $\mathcal{R}(\cdot)$ is pixel-level volume rendering function\cite{mildenhall2021nerf}.
% \begin{align}
%     \mathcal{R}_{\phi_*}(\mathcal{W}_{\phi_w}^{t, \leftarrow}(\mathbf{X}^t))=\sum_{i=1}^{N} \tau_{i} \mathbf{c}\left(\mathcal{W}_{\phi_w}^{t, \leftarrow}\left(\mathbf{X}_{i}^{t}\right)\right).
%     %, o\left(\mathbf{x}^{t}\right)=\sum_{i=1}^{N} \tau_{i}
%     \label{eq:rendering}
% \end{align}
% where the $i$-th sampling point $\mathbf{X^t_i}$ warped to canonical space point $\mathbf{X^*_i}$, $\mathbf{c}$ is pixel color, $\tau_{i}$ is the visible probability of $X_i^t$ computed by the density of warped sampling points $\sigma_i=\sigma(\mathcal{W}_{\phi_w}^{t, \leftarrow}(\mathbf{X}^t_i))$, as defined in \cref{eq:nerf}. 
\begin{figure}[t]
  \centering
  \vspace{-1.3 em}
   \includegraphics[width=1\linewidth]{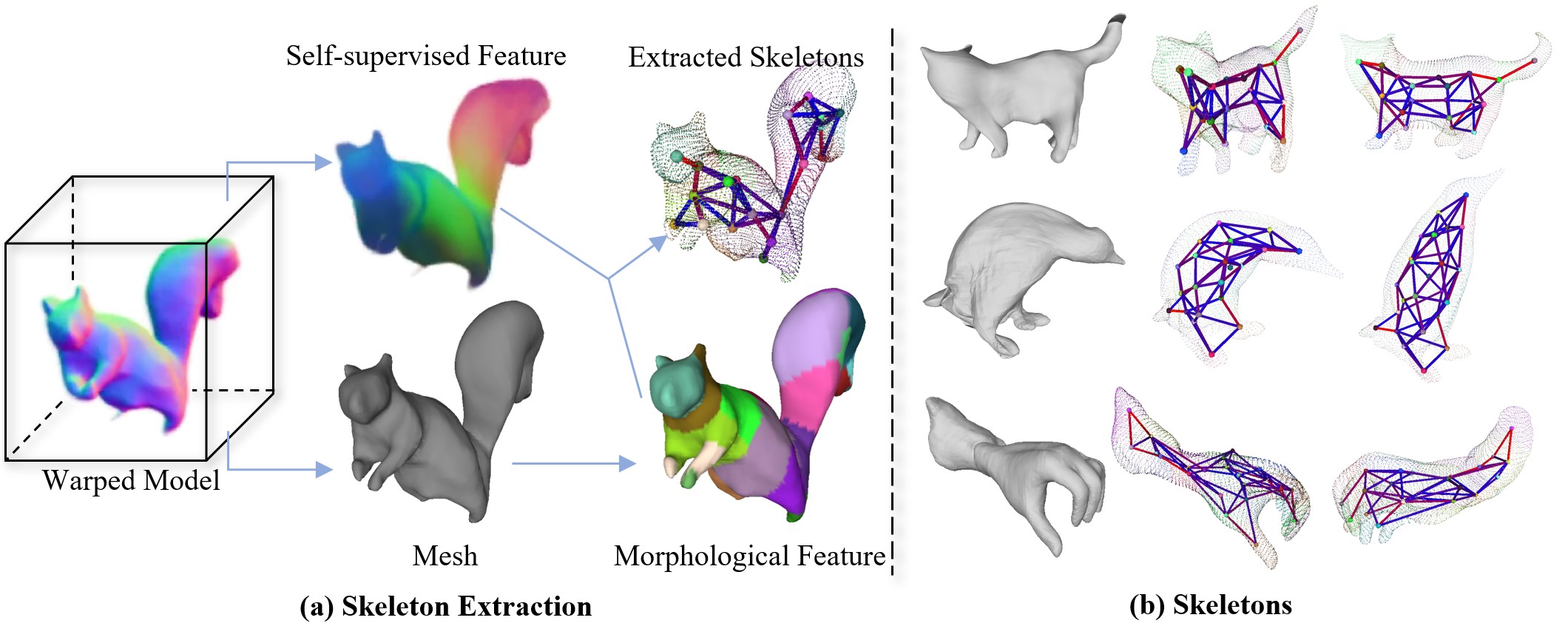}
   \vspace{-1.3 em}
   \caption{\textbf{Skeleton Construction.} (a) Pipeline of skeleton construction. We extract mesh from canonical space and construct skeletons using both semantic correlation and morphological correlation. (b) Constructed skeletons for cat, penguin, and hand.}
   \label{fig:skel}
   \vspace{-2 em}
\end{figure}
%-------------------------------------------------------
\vspace{-1 em}
\subsection{Skeleton-based Generation}
\label{sec:skeleton}
With a deformable model with the bone motions extracted from the reference video, a vanilla way to generate a new text-guided model is to directly modify the canonical model using SDS. However, this raises two problems: first, considering that SDS is an incremental process without a target function, brutally applying SDS to a well-reconstructed model will limit the generated model's diversity. Secondly, the distribution of rendered images from the reconstructed model is different from that of the diffusion model's training set. The reconstructed model may hurt the performance instead. In this context, we proposed skeleton-based generation, which generates from a new start while being constrained by previously extracted motion.

\noindent\textbf{Skeleton Construction.} We pass model irrelevant parameters including the position embedding and time embedding from the reconstructed model to the generated model. Subsequently, we initialize the density field of canonical space of the generated model through bones and skeletons. To extract the skeletons from the reconstructed model, we first extract canonical mesh with edges $\textit{E}=\{\mathbf{e}_i=\{\mathbf{v}_{m}, \mathbf{v}_{n}\}\}$ and vertices $\textit{V}=\{\mathbf{v}_i \in \mathbb{R}^3\}$ using marching cube. Then we estimate the relation of each pair of bones based on the skinning weights and feature descriptor $\psi$ of vertices. The feature descriptor of each bone $b$ is defined as the weighted sum of related vertices in \cref{eq:bone_feat}: 
\vspace{-0.5 em}
\begin{align}
    \psi_b = \frac{\sum_{i=1}^{N}S_{b,*}(\mathbf{v}_{i})(\mathbf{MLP}_{\psi}\left(\mathbf{v}_{i}\right))}{\sum_{i=1}^{N}S_{b,*}(\mathbf{v}_{i})}, \label{eq:bone_feat}
\end{align}
and the semantic correlation of two bones $\mathbf{G}_{j,k}$ is calculated as:
\begin{align}
    \mathbf{G}_{j,k} = \text{softmax} \left( \left \langle \psi_j,\psi_k \right \rangle \right), \label{eq:sc}
\end{align}
where $\left \langle \cdot \right \rangle$ is the cosine similarity score. Though bones with similar features are intended to be connected, it may fail when multiple instances share one semantic feature (e.g. arms of a squirrel). To address this issue, we further explore the morphological correlation matrix $\mathbf{M} \in \mathbb{R}^{B \cdot B}$ of each bone pair. For each edge connecting a pair of vertices, we calculate the $\mathbf{M}$ as defined in \cref{eq:bone_morph}:
\vspace{-0.5 em}
\begin{align}
    \mathbf{M} = \frac{\sqrt{\sum_{e_i=\{\mathbf{v}_{m}, \mathbf{v}_{n}\}}^{L} \mathbf{S}_{*}(\mathbf{v}_{m})\mathbf{S}_{*}(\mathbf{v}_{n})^T}}{L}, \label{eq:bone_morph}
\end{align}
where $L$ is the number of edges, $\mathbf{S}_{*}(\cdot) \in \mathbb{R}^{B}$ is the skinning weight matrix in canonical space. The higher the value of $\mathbf{M}$, the greater the degree to which two bones jointly influence the control over the same surface regions. We balance the semantic correlation and morphological correlation of a pair of bones and define the strength of the skeleton $\mathcal{T}_{j,k}$ as:
\vspace{-0.5 em}
\begin{equation}
    \mathcal{T}_{j,k}=
    \begin{cases}
    \mathbf{G}_{j,k} + \alpha \mathbf{M}_{j,k}& \mathbf{M}_{j,k} \geq \xi \\
    0&\mathbf{M}_{j,k} < \xi
    \end{cases},
\end{equation}
where weighting scalar $\alpha$ and threshold $\xi$ are learned in a hierarchical manner. The constructed skeletons are shown in \cref{fig:skel}.

\noindent\textbf{Constrain with Skeletons and Bones.} Given that we have retained only motion-related parameters and discarded the canonical space model, our goal is to ensure that generated objects remain aligned with the motion, and to prevent model collapse during subsequent optimization of the motion. To achieve this, we employ skeletons and bones as constraints in the generation process. We convert $\mathbf{Q}_b^t$ into a rotation quaternion $\mathbf{R}_{b}^t \in SO(3)$ and a translation vector $\mathbf{T}_b^t$. For $j$-th bone and $k$-th bone share a skeleton, we iterate over all time $t$, and compute the range of relative position $\mathbf{T}_{jk}^t=\|\mathbf{T}_j^t-\mathbf{T}_k^t\|_2$ and the range of quaternion angle $\mathbf{A}_{jk}^t=\angle(\mathbf{R}_{j}^t, \mathbf{R}_{k}^t)$.
% \begin{align}
%     \mathbf{T}_{min,jk}&=\min_{t}(), \\
%     \mathbf{T}_{max,jk}&=\max_{t}(\|\mathbf{T}_j^t-\mathbf{T}_k^t\|_2), \\
%     \mathbf{A}_{min,jk}&=\min_{t}(), \\
%     \mathbf{A}_{max,jk}&=\max_{t}(\angle(\mathbf{R}_{j}^t, \mathbf{R}_{k}^t)).
%     \label{eq:skel_con}
% \end{align}
With this setting, constraints can be applied to the motion of bones, thus preventing motion divergence when generative loss is applied:
\begin{align}
    \mathcal{L}_{skel}&=\lambda_{T}\mathcal{L}_{skel,T}+\lambda_{A}\mathcal{A}_{skel,T},\\
    \mathcal{L}_{skel,T}&=\sum\nolimits_{j,k} \mathcal{T}_{j,k}\max(\mathbf{T}_{jk} - \mathbf{T}_{max}, \mathbf{T}_{min} - \mathbf{T}_{jk}, 0)^2,\\
    \mathcal{L}_{skel,A}&=\sum\nolimits_{j,k} \mathcal{T}_{j,k}\max(\mathbf{A}_{jk} - \mathbf{A}_{max}, \mathbf{A}_{min} - \mathbf{A}_{jk}, 0)^2,
    \label{eq:loss_skel}
\end{align}
where $\mathbf{T}_{\text{max}}$ and $\mathbf{T}_{\text{min}}$ denote the maximum and minimum values of $\mathbf{T}_{jk}^t$ across time $t$, respectively, while $\mathbf{T}_{\text{max}}$ and $\mathbf{T}_{\text{min}}$ represent the maximum and minimum values of $\mathbf{A}_{jk}^t$ across time $t$, respectively.

Besides, we also use bones to constrain the generated surface as well as skinning weights:
\vspace{-0.5 em}
\begin{align}
    \mathcal{L}_{bone}=\sum_{\mathbf{X}}\mathbf{H}\left(d(\mathbf{X}),d_g(\mathbf{X})\right)+\sum_{t, \mathbf{X}^t}\mathbf{S}_{t}(\mathbf{X}^t)\log(\frac{1}{\mathbf{S}_{t}(\mathbf{X}^t)}),
    \label{eq:loss_bone}
\end{align}
where $\mathbf{H}$ denotes the binary cross entropy, and $d$ and $d_g$ represent the Signed Distance Function (SDF) value of the canonical model and Gaussian bones, respectively. The first term aims to ensure that the surface is closely aligned with the Gaussian bones, taking into account their covariance matrix. It also seeks to maintain consistency with the neural skinning weights, thereby enhancing the convergence of generation. The second term is designed to encourage the sparsity of the skinning weights in order to mitigate potential degradation resulting from the first term as well as the generative loss.
% ----------------------------------------------
\vspace{-1 em}
\subsection{Canonical Score Distillation}
\label{sec:csd}
\begin{figure}[t]
\vspace{-0.6em}
  \centering
   \includegraphics[width=1\linewidth]{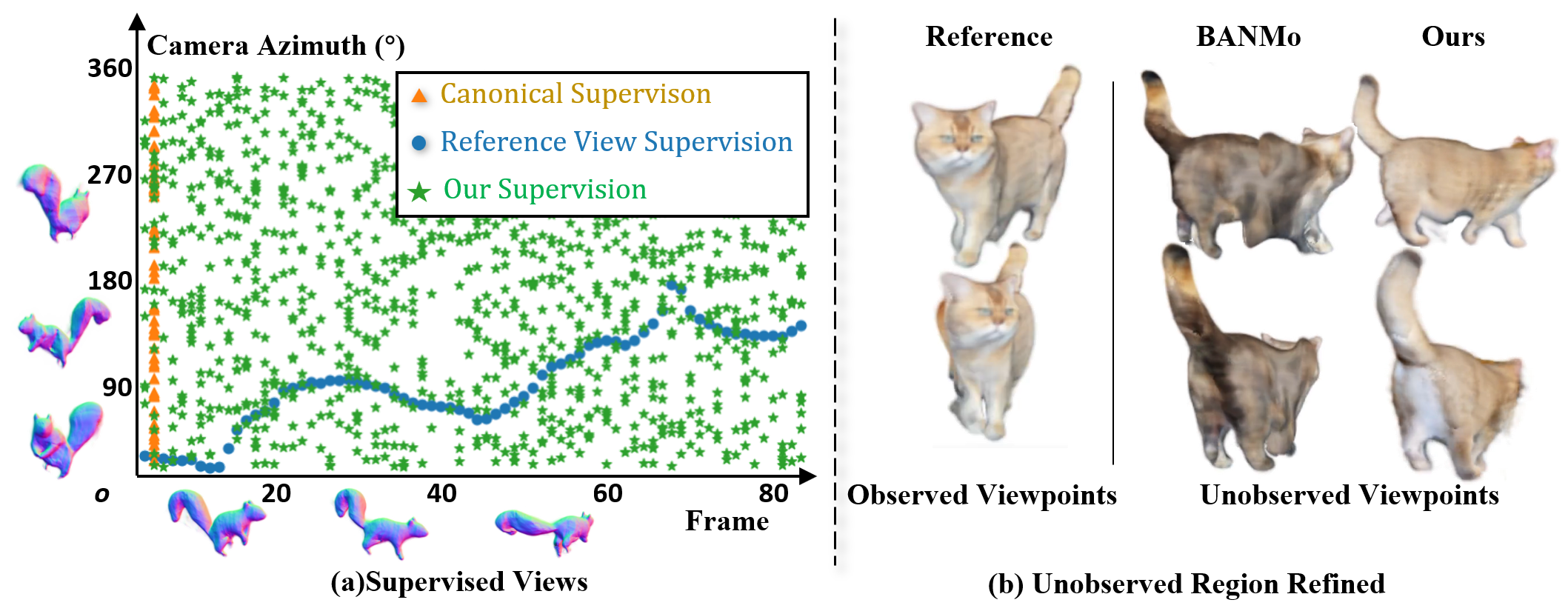}
   \vspace{-1.5 em}
   \caption{\textbf{CSD supervises the entire 4D space.} (a) Our approach can traverse the entire XYZ-t space. In contrast, approaches that focus exclusively on the canonical space or solely on reference views are limited to supervising just a single hyperplane within this 4D space. (b) In unobserved regions, CSD achieves better texture consistency and geometry quality compared to BANMo. }
   \label{fig:attention}
   \vspace{-1.3 em}
\end{figure}

We distillate prior from the diffusion model for bot reconstruction and generation. This 2D supervision is sufficient for static object generation\cite{wang2023prolificdreamer, poole2022dreamfusion}. However, in 4D reconstructing, supervising a single view at a single time point $t$ becomes insufficient. As illustrated in \cref{fig:attention}(a), the supervision from the reference video forms a hyper-plane in the 4D space, resulting in lower quality for unobserved viewpoints. When denoising an image rendered from a viewpoint far away from reference with a large guidance scale, SDS tends to sample a new instance from its own distribution, due to the lack of sufficient information about the original instance.
Here we incorporate multi-view consistent diffusion model MVDream~\cite{shi2023mvdream}, which is trained on a large 2D and 3D dataset~\cite{deitke2023objaverse,schuhmann2022laion}, to generate multi-view consistent images. Cross-view attention spreads the known information to unobserved views, as shown in \cref{fig:attention}(b). 
% By integrating inductive priors from a diffusion model, CSD enhances the authenticity of bones and skinning. With multi-view distillation, CSD infers and reconstructs invisible regions, thereby improving the fidelity of monocular non-rigid generation.
%---------------------------CSD----------------------------------------

\noindent
\textbf{Design of CSD.}
A straightforward way for temporal consistent articulate object generation is to supervise the model in canonical space. However, the canonical model is merely a ``time-slice'' of the 4D object, as depicted in \cref{fig:attention}(a). Roughly supervising the canonical model without considering articulations will result in the degradation of the morphological plausibility of the model in camera spaces. Additionally, unreachable points in canonical space will not be optimized (e.g., occluded body parts at canonical pose), even though they will be rendered in camera space.

To mitigate these issues, we employ a multi-view consistent diffusion model and propose a novel Canonical Score Distillation (CSD) to generate a 4D articulate model with both time consistency and warp robustness across different articulations. We elaborately design CSD to utilize the warping field as a bridge between diffusion prior and canonical model. Conversely, the warp is also refined by diffusion prior, especially in contexts of significant motion and in regions lacking ground-truth images. Distinct from SDS\cite{poole2022dreamfusion}, CSD traverses all frames and replaces the image gradient with two terms: canonical generation and warping refinement, as defined in \cref{eq:CSD}:
% \begin{multline}
%     % \nabla_\phi\mathcal{L}_{CSD}=\mathbb{E}[w(T)(\epsilon_\theta(I_{t,T};y,\mathbf{p},T)-\epsilon)\frac{\partial \mathcal{R}(\phi_*,\mathbf{X}_*)}{\partial \mathbf{X}_*} \frac{\partial W_{\phi_w}(\mathbf{X}_t)}{\partial \phi_w}]
%     \label{eq:CSD}
%         {\footnotesize \nabla_\phi\mathcal{L}_{CSD} (\theta , I_t=g(\phi))=\mathbb{E}_{t,T,\mathbf{p},\epsilon,\mathbf{X}_t}[w(T)(\epsilon_\theta(I_{t,T};y,\mathbf{p},T)-\epsilon)\frac{\partial \mathcal{R}_{\phi_*}(\mathbf{X}_*)}{\partial \mathbf{X}_*} \frac{\partial W_{\phi_w}^{t}(\mathbf{X}_t)}{\partial \phi_w}]}
% \end{multline}
\begin{align}
    \nabla_\phi\mathcal{L}_{CSD}=\mathbb{E}[\underbrace{w(T)(\epsilon_\theta(I_{t,T};y,\mathbf{p},T)-\epsilon)}_{\text{Diffusion Prior}\vphantom{\frac{\partial \mathcal{R}_{\phi_*}(\mathbf{X}_*)}{\partial \mathbf{X}_*}}} \underbrace{\frac{\partial \mathcal{R}_{\phi_*}(\mathbf{X}_*)}{\partial \mathbf{X}_*}}_{\text{\makecell{Canonical\\ Generation}}} \underbrace{\frac{\partial W_{\phi_w}(\mathbf{X}_t)}{\partial \phi_w}}_{\text{\makecell{Warping\\ Refinement}}}]. 
    \label{eq:CSD}
\end{align}

% \vspace{-1.3 cm}
% \begin{center}
% \raisebox{-0.0\height}{\includegraphics[height=4.2\baselineskip]{fig/csd.jpg}}
% \end{center}
% \vspace{-0.2 cm}
 To distinguish from the previous time $t$ of the articulate model, we refer to the time step of diffusion as $T$. $w(T)$ is hyper-parameter controlling the weight of $T$, $y$ is text prompt, $I_t$ are four images rendered in the camera space of time $t$ from four orthogonal view-points $\mathbf{p}$, $I_{t,T}$ are sampled noisy images relative to frame time $t$ and diffusion time step $T$, $\epsilon_\theta(I_{t,T};y,\mathbf{p},T)$ is noise predicted by diffusion model conditioned on prompt and camera poses.
 
\noindent
\textbf{Canonical Generation.} The first term is the gradient of the canonical rendering with respect to the sampling point. It is designed to simplify the generation process from a 4D time-varying model to a series of static 3D models warped from a shared canonical model. Although we traverse all camera spaces, the distillation process is consistently conducted with respect to the canonical model parameters $\phi_*$.

\noindent
\textbf{Warping Refinement.}
Regarding the second term, it signifies that the warping parameters $\phi_w$ are optimized to better collaborate with the canonical model in different poses. As depicted in Fig.~\ref{fig:bones_comp}(a), CSD corrects faulty skinning weights $S_{b,t}(\mathbf{X^t})$ and misplaced bone transformations $\mathbf{Q}^{t}_{b}$.

%%%%%% 冻结正规空间，只优化warpping
%-----------------------------------------
% This design accumulates gradients from different time flows into a same $\phi_*$ and ensures time consistency

% we disentangle deformable object generation into an inductive prior-guided articulation and a diffusion-guided 3D generation, supervising a 4D representation in a spatiotemporally consistent manner.

% \underline{(2)}  (See our videos).
% \textcolor{black}{\underline{(3)} In vanilla SDS, the canonical model parameters are directly exposed to the diffusion prior as $\frac{\partial \mathcal{R}_{\phi_*}(\mathbf{X}_*)}{\partial \phi_*}$. Conversely, $\frac{\partial \mathcal{R}_{\phi_*}(\mathbf{X}_*)}{\partial \mathbf{X}_*}$ in CSD indicates that $\phi_*$ are implicitly optimized through the color gradient corresponding to the warped sampling points $\mathbf{X}_*$. Taking a lower-order derivative with respect to $\phi_*$ can enhance the stability of gradient descent.}
\vspace{-1 em}
\subsection{Optimization}
\label{sec:recon}
% \begin{figure}[t]
%   \centering
%   % \fbox{\rule{0pt}{2in} \rule{0.9\linewidth}{0pt}}
%    \includegraphics[width=0.8\linewidth]{fig/tradeoff.pdf}
%    \vspace{-0.5 em}
%    \caption{\textbf{[X]]Balancing generation and reconstruction trade-off.} This figure illustrates the trade-off between generation and reconstruction. As the loss weights are balanced, the optimal solution shifts from the upper-left corner towards the lower-right. Star-shaped markers indicate points of superior performance compared to utilizing only one type of loss. Notably, ``Generation Error'' is defined as the deviation in model quality from an ideal model.}
%    \label{fig:tradeoff}
%    \vspace{-1.5 em}
% \end{figure}

% \begin{figure}[t]
% \vspace{-0.3em}
%   \centering
%   % \fbox{\rule{0pt}{2in} \rule{0.9\linewidth}{0pt}}
%    \includegraphics[width=0.9\linewidth]{fig/stage.pdf}
%    \vspace{-1.5 em}
%    \caption{\textbf{[Recon rgb -> skel]Two-stage training progress.} In the first stage, we extract the articulation of the template video with CSD as a regularizer. Then we modify the prompt and gradually anneal the reconstruction loss to generate a new model.}
%    \label{fig:stage}
%    \vspace{-0.5 em}
% \end{figure}
We optimize all learnable parameters during the reconstruction phase. Following the extraction of skeletons from the reference video, we optimize a new implicit articulate model with $\mathbf{MLP}_\phi$ and camera-related parameters discarded. The total loss for skeleton extraction and model generation is defined as \cref{eq:lossfn_r} and \cref{eq:lossfn_g}, respectively:
\vspace{-0.5 em}
\begin{align}
\mathcal{L}_{Ext}=\mathcal{L}_{recon}+\mathcal{L}_{CSD}+\mathcal{L}_{reg},\label{eq:lossfn_r}
\end{align}
\vspace{-2 em}
\begin{align}
\mathcal{L}_{Gen}=\mathcal{L}_{skel}+\mathcal{L}_{bone}+\mathcal{L}_{CSD}+\mathcal{L}_{reg}.\label{eq:lossfn_g}
\end{align}
The reconstruction loss, denoted as $\mathcal{L}_{recon}$, comprises the photometric Mean-Square Error (MSE) $\mathcal{L}_{rgb}$, silhouette reconstruction MSE $\mathcal{L}_{sil}$, and flow loss $\mathcal{L}_{\mathrm{OF}}$ \cite{yang2022banmo}, as defined in \cref{eq:reconloss}:
\vspace{-0.5 em}
\begin{equation}\label{eq:reconloss}
    \begin{gathered}
        \mathcal{L}_{recon}=\lambda_{rgb}\mathcal{L}_{rgb} + \lambda_{geo}\left(\mathcal{L}_{sil}+\mathcal{L}_{\mathrm{OF}}\right),\\
        % \mathcal{L}_{\text {rgb }}=\sum_{\mathbf{x}^{t}}\left\|\mathbf{c}\left(\mathbf{x}^{t}\right)-\hat{\mathbf{c}}\left(\mathrm{x}^{t}\right)\right\|^{2}, \\
        % \mathcal{L}_{\text {sil }}=\sum_{\mathbf{x}^{t}}\left\|\mathbf{o}\left(\mathrm{x}^{t}\right)-\hat{\mathbf{s}}\left(\mathrm{x}^{t}\right)\right\|^{2},\\  
        % \mathcal{L}_{\mathrm{OF}}=\sum_{\mathbf{x}^{t},\left(t+t^{\prime}\right)}\left\|\mathcal{F}\left(\mathrm{x}^{t}, t \rightarrow t^{\prime}\right)-\hat{\mathcal{F}}\left(\mathrm{x}^{t}, t \rightarrow t^{\prime}\right)\right\|^{2},
    \end{gathered}
\end{equation}
where silhouette and optical flow are pre-computed with an off-the-shelf model\cite{yang2023track}. $\lambda_{rgb}$ and $\lambda_{geo}$ are balancing weights. The registration loss $\mathcal{L}_{reg}$ is defined in \cref{eq:regloss}:
\vspace{-0.5 em}
\begin{equation}\label{eq:regloss}
    \begin{gathered}
        \mathcal{L}_{reg}=\mathcal{L}_{match}+\mathcal{L}_{2D-cyc}+\mathcal{L}_{3D-cyc},\\
    \end{gathered}
\end{equation}
where $\mathcal{L}_{\mathrm{match}}$, $\mathcal{L}_{\mathrm{2D-cyc}}$, and $\mathcal{L}_{\mathrm{3D-cyc}}$ represent the loss functions for 3D point feature matching~\cite{yang2022banmo}, 2D cycle consistency~\cite{yang2021viser}, and 3D cycle consistency~\cite{li2021neural}, respectively.

Here, we have designed a two-stage schedule with balanced weights for generation. During the articulation extraction stage, the model is primarily supervised by images, with pre-computed mask, flow, and features. Concurrently, the weight of $\mathcal{L}_{CSD}$ is configured to be low for the complementation of unseen regions, thereby ensuring a gentle adjustment without overpowering the original data. For the generation stage, the position embedding bandwidth is set to max value to guarantee the detail of the generated model.

\section{Experiments}

\label{sec:experiments}
We conducted experiments on generation (\cref{sec:expgen}) and reconstruction (\cref{sec:exprecon}) tasks using the Casual Videos dataset and Animated Objects dataset~\cite{yang2022banmo}. The experiments conducted spanned a diverse array of species, such as squirrels, cats, finches, eagles, humans, hands, and manipulators, among others, to convincingly showcase the capability of our method across generic categories. In the context of the generation task, our method excels in creating spatiotemporally consistent, animatable 3D models with text prompts and a template video. With the proposed CSD method, the generated 4D model exhibits superior performance over existing distillation strategies in terms of temporal consistency and warp robustness, as demonstrated in Table~\ref{tab:gen}. For the reconstruction task, our approach significantly outperforms previous methods, particularly when the number of viewpoints and videos is limited, as shown in Table~\ref{tab:recon}.
\vspace{-0.5 em}
\subsection{Technical Details}
\label{sec:detail}
To enlarge the Casual Videos dataset~\cite{yang2022banmo}, we collect videos containing a single complete instance with large kinesis from the internet. We utilize off-the-shelf models ~\cite{yang2023track, yang2021rigidmask,dinov21}to extract mask, optical flow, and features. We modify the camera distance and near-far plane calculation to avoid the articulated model out of frustum or obstructing the camera. Considering that viewpoints are fixed for reconstruction and randomly selected for generation, we alternate the loss calculation of reconstruction and generation in practice. On a single Nvidia A800 GPU, we sampled 128 pixels from 32 images in reconstruction and rendered four $200\times200$ images for generation. The complete training takes 5 hours for 12000 iterations. Here we adopted a gradient cache technology for saving memories\cite{zhang2022arf}.
\vspace{-1 em}
\subsection{Animatable 3D Model Generation} \label{sec:expgen}
\begin{figure}[t]
  \centering
   \includegraphics[width=0.9\linewidth]{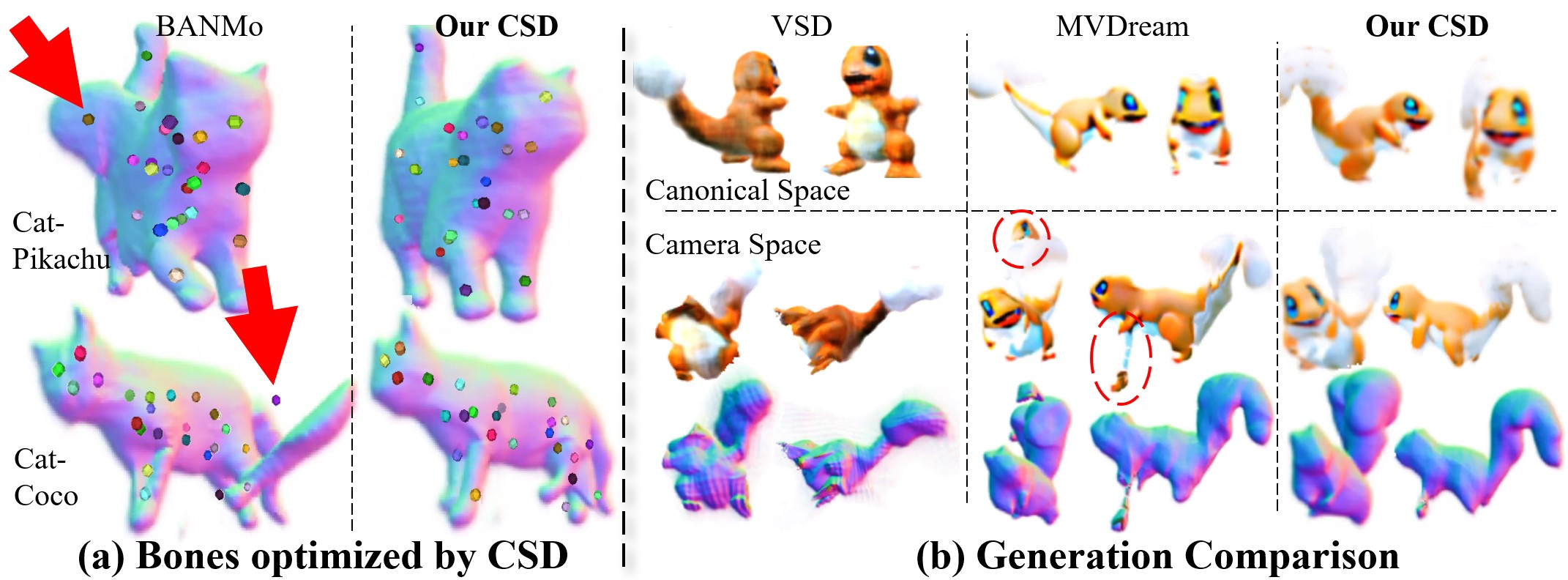}
   \vspace{-0.3 em}
   \caption{\textbf{Canonical score distillation results.} (a) Through the warping refinement term, bones are optimized by the diffusion prior and warped surfaces. (b) Compared with VSD and MVDream, our method generates a more detailed result with plausible motions.}
   \label{fig:bones_comp}
   \vspace{-0.5 em}
\end{figure}

\begin{figure*}[t]
  \centering
   \includegraphics[width=0.9\linewidth]{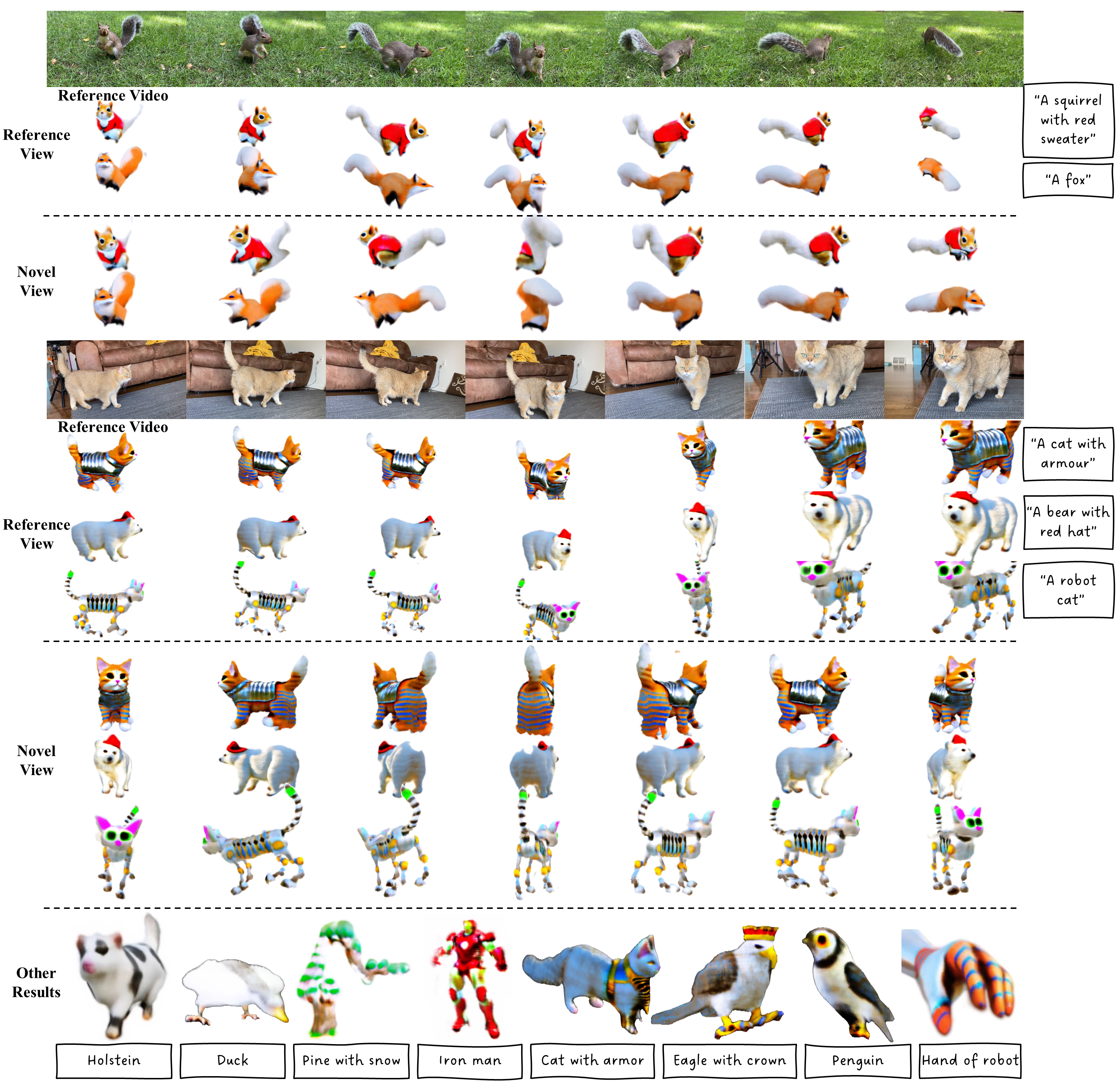}
   \vspace{-0.5 em}
   \caption{\textbf{Text-to-4D Generation.} We generate animatable 3D models by leveraging a text prompt and a template video, achieving diverse outcomes that encompass both texture and geometry while maintaining temporal consistency and morphological plausibility across different poses. See the Appendix for more results.}
   \label{fig:genres}
   \vspace{-0.5 em}
\end{figure*}

\vspace{-0.5 em}
\noindent \textbf{Qualitative Comparisons.} We present our generated results alongside the input videos in \cref{fig:genres}. By disentangling the deformation and canonical model, our generated models demonstrate time consistency, even in cases where the video duration is extensive. Through optimization across all frames, our approach effectively eliminates issues such as disconnected shapes, flickering, and shape inconsistency. Notably, our method is capable of generating various species including quadruped, squirrel, eagle, bird, penguin and so on, and goes beyond mere texture generation with modifying the model's geometry. We employ VSD\cite{wang2023prolificdreamer} and MVDream\cite{shi2023mvdream} as baselines and qualitative comparisons are shown in \cref{fig:bones_comp}(b). Also, we conducted a comparative analysis with several notable 3D generation and 4D reconstruction methods: ProlificDreamer~\cite{wang2023prolificdreamer}, 3D reconstructor BANMo~\cite{yang2022banmo}, texture-swap articulated representation Farm3D~\cite{jakab2023farm3d} and text to video generator Text2Video-Zero~\cite{khachatryan2023text2video} and summarize our strengths in \cref{tab:comp}.
\begin{table}[t]
  \centering
  \scriptsize
  % \vspace{-0.6 em}
  \begin{tabular}{lllllll}%{@{}lc@{}}
    \toprule
    Method           & Prompt     & Category & Generation       & 3D Model & Motion       & Articulation \\
    \midrule
    ProlificDreamer\cite{wang2023prolificdreamer} & Text       & Generic  & Shape+Texture & NeRF & -            & -            \\
    % Hex-Plane\cite{cao2023hexplane}        & -          & Generic  & -                & NeRF & Learned      & -            \\
    Text2Video-Zero\cite{khachatryan2023text2video}  & Text       & Generic  & -                & -        & Learned            & -            \\
    BANMo\cite{yang2022banmo}            & -          & Generic  & -                & NeRF & Learned      & Learned      \\
    Farm3D\cite{jakab2023farm3d}         & Image      & Specific & Texture          & Mesh     & Manual & Pre-defined  \\
    \textbf{Ours}             & \textbf{Video+Text} & \textbf{Generic}  & \textbf{Shape+Texture} & \textbf{NeRF} & \textbf{Learned}      & \textbf{Learned}   \\
    \bottomrule
  \end{tabular}
  \vspace{0.6 em}
  \caption{\textbf{Related work overview on non-rigid 3D model reconstruction and generation.} Distinguishing from previous work, our method, AnimatableDreamer, generates text-guided animatable models across generic categories without the need for pre-defined templates. This attribute establishes AnimatableDreamer as a versatile and user-friendly 4D generation tool.}
  \vspace{-1em}
  \label{tab:comp}
\end{table}

\noindent \textbf{Quantitative Comparisons.} We perform a quantitative evaluation of the generation quality and the consistency between text and images by utilizing the CLIP-score\cite{park2021benchmark} and the GPT-4 Eval3D\cite{wu2024gpt} methodologies. A video is rendered employing the camera trajectory specified in Eval3D\cite{wu2024gpt}, which navigates around the scene at a constant elevation angle while varying the azimuth. Each video frame is then assessed using the CLIP ViT-B/32 model, and the scores are aggregated across all frames and text prompts to compute the overall CLIP score. Additionally, we assess the temporal consistency of CLIP-T by calculating the CLIP similarity between consecutive frames. We employ VSD\cite{wang2023prolificdreamer} and MVDream\cite{shi2023mvdream} as baselines, as presented in \cref{tab:gen}.
\begin{figure*}[t]
  \centering
   \includegraphics[width=0.9\linewidth]{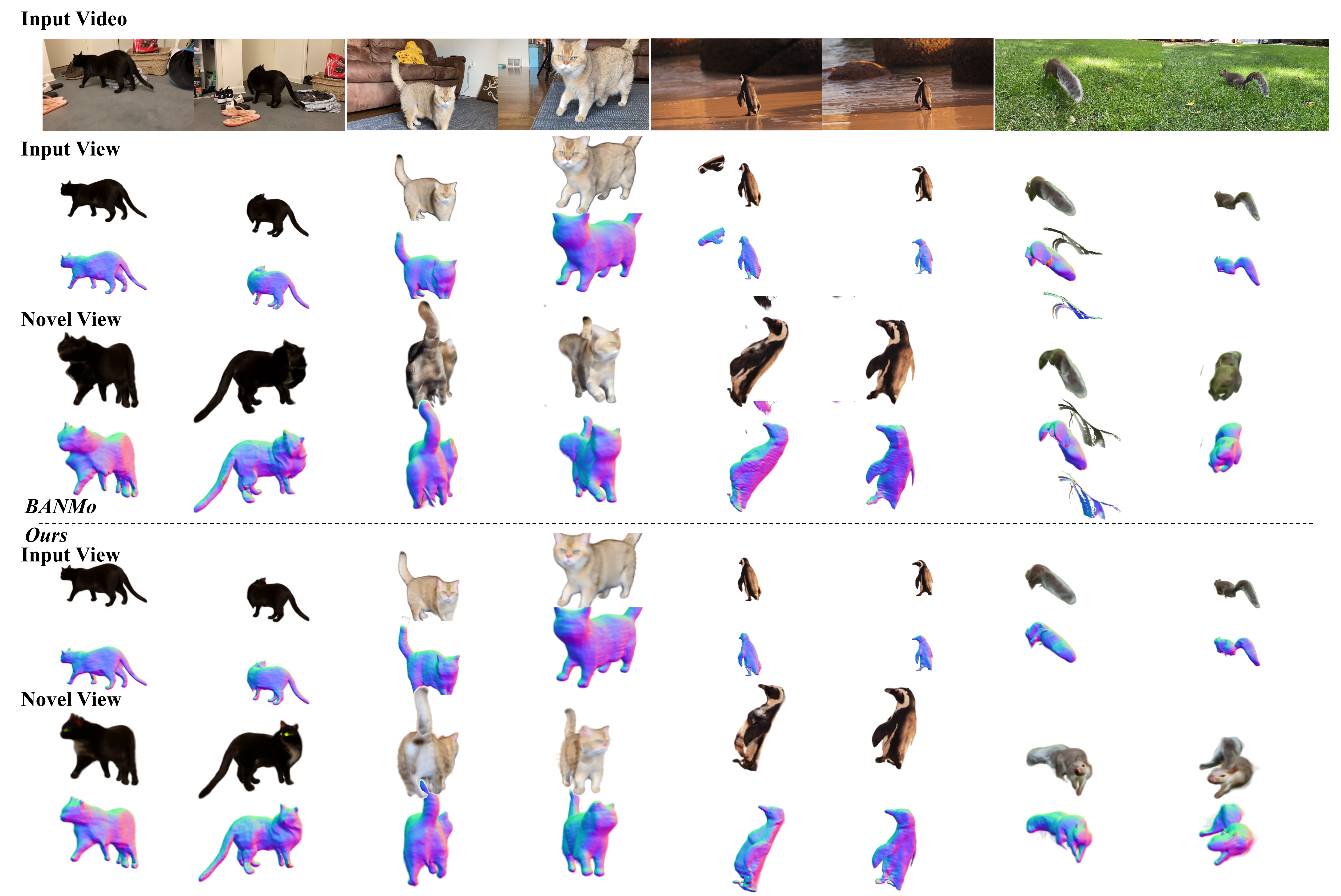}
   \vspace{-0.5 em}
   \caption{\textbf{Monocular reconstruction result on Casual videos dataset.} Our method is visibly superior to BANMo, particularly in frames with large motion (squirrel) or in regions not present in the reference image (cat and penguin). Each experiment is performed based on a \textbf{single monocular video}. See the Appendix for more results.}
   \vspace{-0.5 em}
   \label{fig:reconres}
\end{figure*}
% \begin{table*}[!t]
%     \centering
%     \footnotesize
%     \setlength\tabcolsep{7pt}
%     \label{tab:recon}
%     \begin{tabular}{c|cc|cc|cc|cc|cc}
%       \toprule
%       \multirow{2}{*}{Methods} & \multicolumn{2}{c}{Cat-Coco} & \multicolumn{2}{c}{Cat-Pikachu} & \multicolumn{2}{c}{Penguin} & \multicolumn{2}{c}{Shiba} & \multicolumn{2}{c}{Eagle}\\
%       \cline{2-3}\cline{4-5}\cline{6-7}\cline{8-9}\cline{10-11}
%       & CD  & F@$\%2$ & CD  & F@$\%2$  & CD  & F@$\%2$  & CD  & F@$\%2$ & CD & F@$\%2$  \\
%       \midrule
%       BANMo\cite{yang2022banmo}  &10.7& 15.3& 3.7 &57.3 & 6.4&43.9 &6.81 &36.6 &16.0& 18.1\\
%       \textbf{Ours}  &\textbf{3.65}& \textbf{63.3}& \textbf{2.0} &\textbf{88.9} & \textbf{3.7}&\textbf{64.0} &\textbf{4.54} &\textbf{53.9} &\textbf{12.8}& \textbf{21.3}\\
%       \bottomrule
%     \end{tabular}
%     \vspace{-0.5 em}
%     \caption{\textbf{Quantitative results of monocular reconstruction on Casual Videos and Animated Objects.} We calculate the Chamfer distance (cm, $\downarrow$) and F-score ($\%$, $\uparrow$), averaging the results over all frames and videos.}
%     \vspace{-0.5 em}
% \end{table*}
\begin{table}[!t]
    \centering
    \small
    \begin{tabular}{c|cccc}
      \toprule
      Methods & CLIP$\uparrow$  & CLIP-T$\uparrow$ &R-Precision@10$\uparrow$ & GPT Eval3D elo $\uparrow$ \\
      \midrule
      ProlificDreamer\cite{wang2023prolificdreamer}  &33.1& 95.9&56.3& 959\\ 
      MVDream\cite{shi2023mvdream} &34.8& 94.4&31.2&  979\\
      \hline
      w/o $\mathcal{L}_{bone/skel}$ &27.1& 94.2&35.6&  954\\
      w/o $\mathcal{L}_{skel}$ &28.4& 94.0&40.1&  960\\
      w/o $\mathcal{L}_{bone}$ &37.8& 96.1&81.7&  1070\\
      \textbf{Our CSD}  &\cellcolor{yellow}\textbf{38.2}& \cellcolor{yellow}\textbf{96.6}& \cellcolor{yellow}\textbf{87.5} & \cellcolor{yellow}\textbf{1098}\\
      \bottomrule
    \end{tabular}
    \vspace{0.6 em}
    \caption{\textbf{Quantitative results for generation.} We use CLIP ViT-B/32 for evaluation. CLIP-T is the average CLIP between adjacent frames, CLIP R-Precision@10 is a metric to evaluate the text-image consistency. GPT Eval3D\cite{wu2024gpt} is an evaluator for Text-to-3D generation. Our method outperforms previous works, and we find that without skeleton restriction, the generation tends to diverge.} %
    \vspace{-1 em}
    \label{tab:gen}
\end{table}
\begin{table}[!t]
    \centering
    \small
    \setlength\tabcolsep{7pt}
    \resizebox{1\linewidth}{!}{
    \begin{tabular}{c|c|cc|cc|cc|cc}
      \toprule
      \multirow{2}{*}{Methods} & \multirow{2}{*}{\makecell{Number\\of videos}} & \multicolumn{2}{c}{Cat-Coco} & \multicolumn{2}{c}{Cat-Pikachu} & \multicolumn{2}{c}{Penguin} & \multicolumn{2}{c}{Shiba} \\
      \cline{3-10}
      & & CD$\downarrow$  & F@$\%2$$\uparrow$ & CD$\downarrow$  & F@$\%2$$\uparrow$  & CD$\downarrow$  & F@$\%2$$\uparrow$  & CD$\downarrow$  & F@$\%2$$\uparrow$ \\
      \midrule
      BANMo\cite{yang2022banmo} &1  &10.7& 15.3& 3.71 &57.3 & 6.47&43.9 &6.81 &36.6\\
      BANMo\cite{yang2022banmo} &4  &4.66& 51.6& 4.51 &52.7 & 3.75   & 60.3    &4.66 &51.9 \\
      RAC\cite{yang2023reconstructing} &1  & 6.25&42.2 &3.60 &60.2 & 4.68 &53.7 & 7.94&30.1 \\
      RAC\cite{yang2023reconstructing} &4  & 4.48&55.8 &3.39 &68.1 & 8.77 &26.9 &6.86 &41.9 \\
      \hline
      w/o $\mathcal{L}_{CSD}$ &1  &8.34& 32.6& 3.88 &59.6 & 6.94&42.3 &5.83 &35.2\\
      \textbf{Ours} &\cellcolor{yellow}\textbf{1}& \cellcolor{yellow}\textbf{3.65}& \cellcolor{yellow}\textbf{63.3}& \cellcolor{yellow}\textbf{2.0} &\cellcolor{yellow}\textbf{88.9} & \cellcolor{yellow}\textbf{3.7}&\cellcolor{yellow}\textbf{64.0} &\cellcolor{yellow}\textbf{4.54} &\cellcolor{yellow}\textbf{53.9} \\
      \bottomrule
    \end{tabular}
    }
    \vspace{0.6 em}
    \caption{\textbf{Quantitative results of monocular reconstruction on Casual Videos and Animated Objects.} We calculate the Chamfer distance (cm, $\downarrow$) and F-score ($\%$, $\uparrow$), averaging the results over all frames and videos. Leveraging prior from diffusion, our method outperforms existing methods, despite the absence of multiple videos or templates.}
        \label{tab:recon}
    \vspace{-2em}
    
\end{table}
% ------------------------------
%\multirow{-2}{*}{\centering\arraybackslash}
\vspace{-1.5 em}
\subsection{Animatable 3D Model Reconstruction}
\vspace{-0.5 em}
\label{sec:exprecon}
\textbf{Qualitative Comparisons.} As depicted in \cref{fig:reconres}, we outperform the existing animatable object reconstruction method on Casual Videos dataset and Animation Objects dataset (\cref{sec:experiments}). A notable observation is that our approach goes beyond merely refining the canonical model. It also encompasses modifications to the bones and warping fields, as stated in \cref{eq:CSD}. This comprehensive modification is especially effective in scenarios where the initial bone structures are implausible. In such cases, our method adeptly repositions the bones, accompanied by corresponding adjustments to the warping function, which is clearly evidenced in \cref{fig:bones_comp}(a). It is evident that while BANMo delivers good results at the input views, its performance significantly diminishes in unobserved spaces. This discrepancy in quality can be attributed to the model's tendency to overfit on the reference views, especially when there is a lack of supervision from other viewpoints, as we discussed in \cref{fig:attention}. 

%For quantitative comparisons of reconstruction, we compare our method with BANMo~\cite{yang2022banmo} on Animated Objects dataset and Casual Videos dataset~\cite{yang2022banmo}
\noindent \textbf{Quantitative Comparisons.} Our evaluation utilizes both Chamfer distances~\cite{tatarchenko2019single} and F-scores using a threshold set to 2\% of the bounding box size. Considering that Casual Videos has no ground truth, we employ BANMo on multiple video sequences and extract meshes to serve as pseudo ground truth. The result in \cref{tab:recon} suggests {AnimatableDreamer} significantly outperforms BANMo, even when the pseudo ground truth is derived from BANMo itself. This indicates that, with CSD, our framework effectively supplements the missing information, surpassing the need for additional video data.
\vspace{-1em}
\subsection{Ablation Study} For the generation process, we conduct ablation studies on $\mathcal{L}_{bone}$ and $\mathcal{L}_{skel}$ as detailed in \cref{tab:gen}. Our findings indicate that the absence of skeletal constraints $\mathcal{L}_{skel}$ leads to divergence in generation or results in motions becoming disconnected from the model. Additionally, it is observed that incorporating $\mathcal{L}_{bone}$ enhances the surface quality of the generated models. In the context of reconstruction, the ablation of $\mathcal{L}_{CSD}$ reveals a significant enhancement in performance, for it refines the texture and geometry of unobserved regions. Please refer to the Appendix for more results.
% \vspace{-1.5em}
\section{Conclusion}
% \vspace{-1em}
In this work, we present {AnimatableDreamer}, a pioneering framework for the generation and reconstruction of generic-category non-rigid 3D models. With the proposed Canonical Score Distillation (CSD), {AnimatableDreamer} addresses the challenges of unconstrained deformable object generation by simplifying the 4D generation problem into 3D space. Our method excels in generating diverse spatial-temporally consistent non-rigid 3D models based on textual prompts. With articulations extracted from monocular video, users can manipulate and animate these models by controlling the rigid transformations of bones. We demonstrate improved performance compared with existing monocular non-rigid body reconstruction methods, especially in scenarios with limited viewpoints and substantial motion.

\noindent \textbf{Limitations.} Our method requires large VRAM to render high-resolution images for CSD during training, due to the long gradient chain from camera space to canonical space, posing constraints on the optimization process. The requirement to feed four images into MVDream simultaneously poses a computational burden further. Addressing these issues could enhance the overall performance and versatility. 


\bibliographystyle{splncs04}
\bibliography{main}
\end{document}